\documentclass[letterpaper, 10 pt, conference]{ieeeconf}  

\IEEEoverridecommandlockouts                              

\usepackage{hyperref} 
\usepackage{mathrsfs}
\usepackage{amsfonts}
\usepackage{amsmath}
\usepackage[flushleft]{threeparttable}
\usepackage{tablefootnote}
\usepackage{multirow}
\usepackage{graphicx}
\usepackage{subfigure}
\usepackage{hanging}
\usepackage{color}
\usepackage{tikz}
\usetikzlibrary{calc,arrows,decorations.markings}
\usepackage{balance}
\usepackage{cite}
\usepackage{xpatch}
\makeatletter
\patchcmd\@makecaption{\\}{.~}{}{\fail}
\makeatletter

\title{\LARGE \bf
Relationship Oriented Affordance Learning through \\ Manipulation Graph Construction
}

\author{Chao Tang$^{1}$, Jingwen Yu$^{1,2}$, Weinan Chen$^{1}$, \emph{Member, IEEE}, and Hong Zhang$^{1}$ \emph{Fellow, IEEE}
\thanks{$^{1}$Department of Electronic and Electrical Engineering, Southern University of Science and Technology, China.}%
\thanks{$^{2}$Department of Electronic and Computer Engineering, Hong Kong University of Science and Technology, Hong Kong, China.}%
}

\begin{document}

\maketitle
\thispagestyle{empty}
\pagestyle{empty}


\begin{abstract}

In this paper, we propose Manipulation Relationship Graph (MRG), a novel affordance representation which captures the underlying manipulation relationships of an arbitrary scene. To construct such a graph from raw visual observations, a deep nerual network named \emph{AR-Net} is introduced. It consists of an \emph{Attribute} module and a \emph{Context} module, which guide the relationship learning at object and subgraph level respectively. We quantitatively validate our method on a novel manipulation relationship dataset named SMRD. To evaluate the performance of the proposed model and representation, both visual perception and physical manipulation experiments are conducted. Overall, \emph{\textbf{AR-Net}} along with MRG outperforms all baselines, achieving the success rate of 88.89\% on task relationship recognition (TRR) and 73.33\% on task completion (TC). 

\end{abstract}

\section{INTRODUCTION}

Humans possess rich prior knowledge as for recognizing what function or interaction an object affords in order to accomplish a certain task. Similarly, robots operating in the unconstructed environment ought to be equipped with the ability to reason about object affordances to plan task-oriented manipulation sequences. Affordance, first mentioned in \cite{gibson1977theory}, enables robots to identify its potential interactions with humans or environments and informs the subsequent robotic manipulation \cite{xu2021affordance, qin2020keto, daruna2019robocse}. It places humans and robots on the same page in terms of cognitive reasoning and scene understanding. This paper is concerned with the following question: given an arbitrary scene consisting of a set of objects, how does the robot figure out what to do with them. For example, a cooking robot should be able to recognize a spatula can be used to scoop the pot or handover appropriate kitchen tools to a human chef during cooking. Due to the importance of affordance learning, extensive research has been conducted.

Most works in computer vision community cast affordance learning as a segmentation problem \cite{myers2015affordance, do2018affordancenet, chu2019recognizing, deng20213d}. Object parts sharing the same functionality are segmented and grouped at the pixel level before assigned with corresponding affordance category labels. These methods treat affordance learning as detecting the property of an isolated object and do not consider the interactions among objects. For example, a robot predicts the affordance of a knife as ``cut'' (blade) and ``grasp'' (handle) but has no idea as to ``what to cut'' or ``where to cut''. Contrary to the segmentation based approach, another approach known as knowledge graph relies on a symbolic representation of a scene where detected objects are first described in an embedding space. Then a knowledge graph is constructed whose nodes denote a set of concepts known to the robot (e.g. apple, mug, bowl) and whose edges denote a set of possible relations such as affordances \cite{daruna2019robocse, saxena2014robobrain, beetz2018know, daruna2021continual, liu2004conceptnet}. Now, to query the learned knowledge graph, the robot must recognize the object category first using pretrained object detectors such as \cite{ren2015faster}, which makes the closed world assumption, and is unable to generalize with respect to novel object instances.

\begin{figure}[t]
  \centering
  \begin{tikzpicture}[inner sep = 0pt, outer sep = 0pt]
    \node[anchor=south west] (fnC) at (0in,0in)
      {\includegraphics[height=2.3in,clip=true,trim=0in 0in 0in 0.5in]{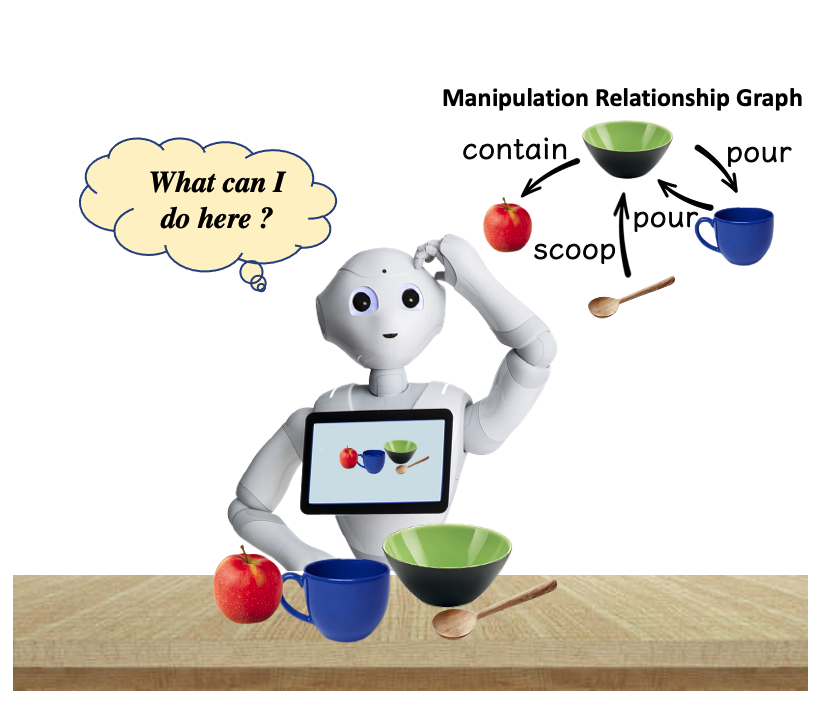}};
  \end{tikzpicture}
  \vspace*{-0.1in}
  \caption{Robot reasons about the manipulation relationship graph (MRG) over a set of kitchen objects.} 
  \label{fig:concept}
  \vspace*{-0.25in}
\end{figure}

To extend the previous works in affordance learning, we introduce manipulation relationship graph (MRG), a novel representation which interprets affordance as the manipulation relationships between pairs of objects, and show how an MRG is constructed from raw visual observations without object category information. Rather than asking the question of what functionality an object is able to afford or provide, our affordance framework addresses the question of how objects in a scene are related in terms of functionalities, i.e., how pairs of objects are related to each other through affordance properties. As is depicted in Figure \ref{fig:concept}, when the robot first perceives a scene consisting of a set of objects, it abstracts the scene to an MRG whose nodes denote the object observations and edges indicate the potential manipulation relationships between pairs of objects. In general, each manipulation relationship is written as a triplet ($subject$, $relationship$, $object$), where $subject$ and $object$ are represented by an index (e.g. with four objects, indexes runs from 1 to 4). Such an MRG guides robotic manipulation by informing ``what to use" (subject), ``how to use"(relationship) and ``what to act on" (object). 

To construct the MRG, we contribute \emph{\textbf{AR-Net}}, a novel affordance learning framework which captures the underlying manipulation relationships of an arbitrary scene. It consists of an \emph{Attribute} module and a \emph{Context} module to facilitate model's generalization to intra-class and inter-class object variations. The manipulation relationships are learnt under the supervision of a novel dataset, which will be introduced in detail shortly. In this paper, we use the word \textbf{object} to either refer to the physical entities being manipulated or the recipient of an action in a ($subject$, $relationship$, $object$) triplet. Which interpretation is relevant should be clear from the context. Our contributions in this paper are three-fold:

\begin{itemize}
    \item We introduce MRG, a novel affordance representation which captures the underlying manipulation relationships between pairs of objects directly from raw visual observations.
    \item We propose \emph{\textbf{AR-Net}}, a deep neural network to infer the manipulation relationships and construct the MRG.
    \item A novel dataset, semantic manipulation relationship detection (SMRD) dataset, consisting of six manipulation relationship classes, 13 tool categories, 1171 images, and 4428 relationship triplets, is presented.
\end{itemize}

The rest of this paper is organized as follows. We review related literature in Section \ref{related_works} and provide the formal definition of MRG in Section \ref{MRG}. The detailed explanation of the proposed \emph{\textbf{AR-Net}} is given in Section \ref{approach}. Experimental setup and results for both visual perception and physical manipulation experiments are presented in Section \ref{exp_setup} and \ref{exp} respectively. We discuss the limitations and provide a conclusion in Section \ref{discussion}.

\section{Related Works} \label{related_works}

In this paper, we propose to learn affordance by constructing MRG. With respect to prior research, three most related topics are reviewed in this section including affordance learning in robotics, visual relationship detection, and manipulation using a graph.

\subsection{Affordance Learning in Robotics}

Affordance provides valuable information as to what function an object or environment affords for the purpose of achieving a specified goal by robots \cite{gibson1977theory}. There are currently three affordance learning approaches: segmentation based, knowledge-graph based and interaction based, each of which presents a unique way to interpret affordance. The segmentation based approach segments the object part by functionality at the pixel level. \cite{myers2015affordance} starts with segmenting tool affordance regions from local shape and geometry primitives. Built on prior works, \cite{do2018affordancenet} utilizes deep learning to achieve state-of-the-art performance on affordance segmentation. While the aforementioned methods operate on 2D or 2.5D data, \cite{deng20213d} segments objects in the 3D domain. The segmentation based approach usually focuses on detecting the properties of isolated objects and does not consider the interactions among them. 

Another way to learn affordance is through the knowledge graph framework, which captures the relationships among symbols in an embedding space. \cite{daruna2019robocse} learns multi-relational embeddings to jointly reason about object affordances, locations, and materials. To keep the knowledge base up-to-date, \cite{daruna2021continual} proposes an incremental robotics knowledge graph framework by leveraging continual learning. These methods are usually trained on large knowledge sources such as \cite{beetz2018know} \cite{liu2004conceptnet}. A knowledge-graph based method models affordance as a set of relationships among objects but relies on a pretrained object detector to convert raw visual observations to a set of symbols and is therefore unable to generalize to novel object instances. 

The third affordance learning approach discovers affordances through physical interactions between robots and objects. \cite{qin2020keto} and \cite{turpin2021gift} learn affordances from goal-directed tool manipulation via a set of predicted keypoints in simulation. Similarly, \cite{xu2020deep} reasons about the long-term effects of actions through modeling what actions are afforded in the future through trial-and-error. These models are usually task-specific and are not able to handle general situations. In contrast to the three affordance learning approaches, our proposed method deals with various tasks and is in principle capable of generalizing to any arbitrary scene without the need for object category information, which is particularly helpful in human robot interaction.

\subsection{Visual Relationship Detection}
Visual relationship detection is defined as the task of describing the relationships between objects within a scene using visual input. \cite{johnson2015image} first proposes to represent objects, attributes and relationships between objects using a scene graph. \cite{lu2016visual} improves on \cite{johnson2015image} by leveraging language priors from semantic word embedding to fine-tune the likelihood of a predicted relationship. After that, numerous methods have been proposed by utilizing recurrent architecture \cite{xu2017scene}, attention mechanism \cite{li2018factorizable} and reinforcement learning \cite{liang2017deep}.

In robotics, \cite{yang2017support} detects the support relationships between surface from RGBD data for the purpose of robot navigation while \cite{zampogiannis2015learning} detects the evolution of the spatial relationships between involved objects during manipulation.Our proposed method extends the task of \emph{describing} the scene with spatial or action relationships to \emph{predicting} potential manipulation relationships among objects. 

\subsection{Manipulation Using A Graph}
Since robotic manipulation necessitates robot executing a sequence of actions in order to achieve a goal, graph becomes a natural spatio-temporal representation. \cite{zhang2018visual} and \cite{park2020single} focus on the problem of robotic grasping. They simultaneously detect objects and recognize the low-level manipulation relationships as a graph to help robots find the right order in which the objects should be grasped. 

Graph has also been widely used in task-and-motion planning (TAMP). \cite{simeonov2020long} solves the long-horizon planning problem by representing subgoals as graphs for rigid-body manipulation. Rather than focusing on symbolic information, \cite{driess2021learning} jointly predicts a sequence of discrete actions and the parameter values of their associated low-level controllers without prior geometric knowledge of the environment. Combining symbolic and geometric reasoning, \cite{zhu2020hierarchical} builds a two-level scene graph representation including geometric and symbolic scene graph for long-horizon manipulation tasks. Following the practice, we use the form of graph to capture the underlying manipulation relationships, but the proposed method focuses on high level scene understanding and uses a graph to capture the diverse manipulation relationships.

\section{Manipulation Relationship Graph} \label{MRG}
\begin{figure}[h]
  \centering
  \begin{tikzpicture}[inner sep = 0pt, outer sep = 0pt]
    \node[anchor=south west] (fnC) at (-0in,0in)
      {\includegraphics[height=1.46in,clip=true,trim=0in 0in 0in 0in]{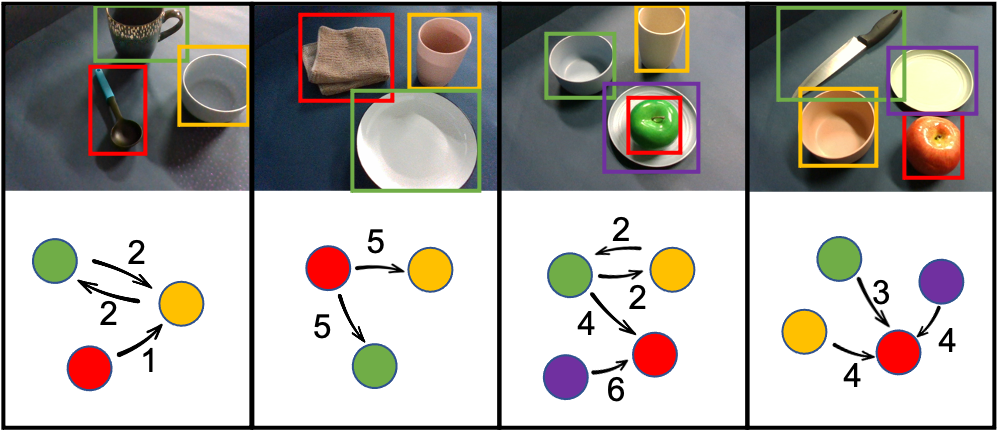}};
  \end{tikzpicture}
  \vspace*{-0.25in}
  \caption{Examples of scene (first row) and corresponding MRGs (second row). In our case, $R$ = 6. Indexes are color-coded and manipulation relationships are represented as follow: \{1-scoop, 2-pour, 3-cut, 4-contain, 5-wipe, 6-dump\}.} 
  \label{fig:MRG_example}
\end{figure}

Given the observation of a scene, our objective is to abstract it to a directed graph which captures the underlying manipulation relationships. For each scene, let there be $N$ objects. A manipulation relationship graph $G = (V, E)$ consists of the vertices, $V = \{v_{i}\}_{i=1}^{N}$ to represent the observed $N$ objects in the scene, and edges $E=\{e_{ij}|$ if the manipulation relationship exists from objects $i$ to $j\}$ for $i, j \in {\{1,...,N\}}$. $R$ classes of manipulation relationships are defined. Examples of scene and the corresponding MRGs are shown in Figure \ref{fig:MRG_example}.

Each node $v_i \in V$ in the graph has a feature vector $\phi_i(v)$, which contains the identity, spatial and appearance information. Since category label is not used, object identity is a 1-dimensional index. Spatial information is represented as a 5-dimensional vector indicating the bounding box coordinates and a confidence score. Appearance information is a feature vector extracted from a visual patch at corresponding spatial location. Edge types and weights (i.e. relationship classes and confidence scores) are inferred using $\phi_i(v)$ of all objects.

\begin{figure*}[t]
  \centering
  \begin{tikzpicture}[inner sep = 0pt, outer sep = 0pt]
    \node[anchor=south west] (fnC) at (0in,0in)
      {\includegraphics[height=2.2in,clip=true,trim=0in 0in 0in 0in]{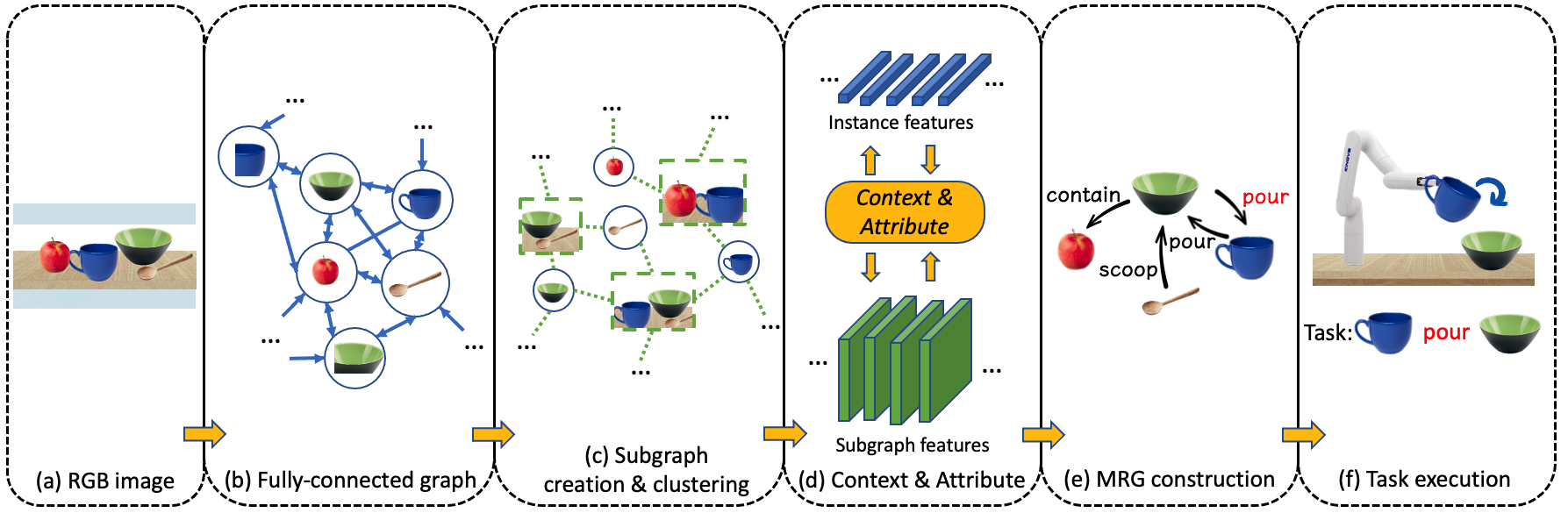}};
  \end{tikzpicture}
        \vspace*{-0.1in}
  \caption{Overview of the proposed \emph{\textbf{AR-Net}}. (a) a RGB image capturing a set of objects is used as input. (b) Objects proposals (blue) generated by RPN are grouped into pairs to construct a fully-connected graph, where each pair is mutually connected by directed edges. (c) Subgraphs (green) are created by taking the union box of two proposals and those referring to the same manipulation relationship are clustered and merged. (d) Object and subgraph features are refined through \emph{Context} module and \emph{Attribute} module. (e) Manipulation relationships are predicted to build MRG. (f) Robot executes the task relationship using computed grasp and effect points.}
  \label{fig:pipeline}
  \vspace*{-0.1in}
\end{figure*}

\section{Relationship Oriented Affordance Learning} \label{approach}
To address the problem of affordance learning in the form of MRG constrcution, we present \emph{\textbf{AR-Net}}, which takes as input the observation of an arbitrary scene and outputs the manipulation relationships among objects as an MRG. An overview of the proposed method is shown in Figure \ref{fig:pipeline}. The entire process can be summarized as follows: (1) given the RGB input, category-agnostic object proposals are grouped into pairs to form a fully-connected graph, where each pair is mutually connected with directed edges; (2) subgraphs are created by taking the union box of two proposals; (3) subgraphs referring to the same manipulation relationship are clustered; (4) object and subgraph features are refined through \emph{Context} and \emph{Attribute} modules of \emph{\textbf{AR-Net}}; (5) manipulation relationships are predicted based on corresponding subject, object and subgraph features to construct an MRG. Each step will be explained in detail for the rest of this section. 

\subsection{FC Graph Construction and Subgraph Clustering}
RPN \cite{ren2015faster} is first adopted to generate category-agnostic object proposals. Without prior knowledge, each pair of objects has possibly mutual manipulation relationships. Therefore, a directed fully-connected graph is constructed as shown in Figure \ref{fig:pipeline} (b), where each object is connected to all the others. Each edge represents a potential manipulation relationship (or no relationship) between subject and object. And each relationship triplet is described as ($subject$, $relationship$, $object$). 

To obtain the representation of a candidate relationship triplet, the union box of subject and object is taken. The corresponding subgraph features are then extracted from the spatial location of the union box. Since many subgraphs share the same subject, object and manipulation relationship, subgraph clustering \cite{li2018factorizable} is deployed to merge these subgraphs (Figure \ref{fig:pipeline} (c)). It helps keep the overall graph representation concise and meanwhile save the computation cost. After clustering, the fully-connected graph consists of two parts: object proposals and subgraphs. Fully-connected layers and 2-D convolution layers are then applied to object and subgraph features respectively to obtain the object feature vectors and 2-D subgraph feature maps. Each subgraph is connected to all the objects within it, and each candidate relationship triplet refers to one subgraph with the corresponding subject and object.

\subsection{Attribute Learning as An Auxiliary Task}
To support the open-world assumption, the object category detection branch is modified to an objectness detection branch. Instead of predicting the category label, it predicts the object existence label as binary classification (``0" for non-object and ``1" for object), making the proposed method category-agnostic. However, removing the category label leads to a considerable drop in relationship prediction performance as it provides important contextual information. 

To resolve the dilemma, \emph{Attribute} module inspired by \cite{chu2019recognizing} is introduced. The main idea is that the key of predicting the category label of either subject or object is to recognize its commonly used functions or attributes. Correctly predicting the attributes seamlessly reduces the searching space of potential manipulation relationships in which the subject or object may be engaged. Meanwhile, attribute is more transferable than category as objects from different categories may share the same attribute. Combining these two leads to better manipulation relationship recognition capability towards novel object instances.

\begin{figure}[t]
  \centering
  \begin{tikzpicture}[inner sep = 0pt, outer sep = 0pt]
    \node[anchor=south west] (fnC) at (0in,0in)
      {\includegraphics[height=1.1in,clip=true,trim=0in 0in 0in 0in]{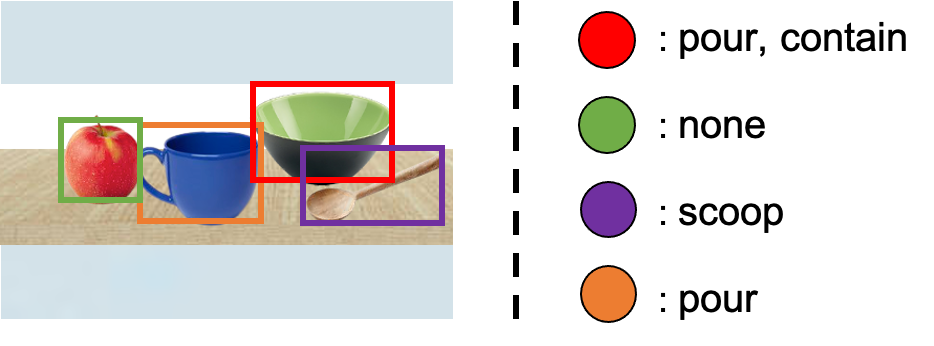}};
  \end{tikzpicture}
  \vspace*{-0.1in}
  \caption{Attribute prediction as an auxiliary task} 
  \label{fig:attribute}
    \vspace*{-0.25in}
\end{figure}

The \emph{Attribute} module explicitly predicts the attributes of a given object as a multi-label classification problem with $R$ outputs, where $R$ is the number of manipulation relationship classes (in our case, $R$=6). As shown in Figure \ref{fig:attribute}, bowl possesses the attributes of ``pour" and ``contain" as humans usually put food in it or use bowl to transfer liquid to other containers. Spoon is solely used for scooping, so it has single attribute. Based on the fact that visually similar parts share similar functions, \emph{Attribute} module guides object feature learning by connecting the visual structure to attributes. It helps the proposed method recognize attributes of novel objects and therefore correctly predict the manipulation relationships. Attribute learning works as an auxiliary task during training and is dropped during inference. Non-affordance objects (e.g. apple) are not considered. The overall detection branch consists of an objectness detection branch, a bounding box regression branch and an attribute prediction branch. 

Attribute learning can be interpreted as novel object detection, which is still an open problem. In the current setup, the model is studied in the limited context of six commonly used attributes among kitchen tools. Ideally, extra attribute categories can be incorporated in order to handle more diverse and challenging scenarios. Different from adding many more tool categories, adding one more attribute category may potentially enable the robot to reason about the usage of a large set of tool categories with similar functions.

\subsection{Subgraph Feature Aggregation with Context}
Object features offer detailed contextual cues such as appearance and spatial location information to facilitate the understanding of how each candidate relationship region (i.e. subgraph) interacts with its corresponding subject and object. In other words, a candidate relationship region might contain several objects due to dense layout or inaccurate bounding box localization. But the manipulation relationship is only determined by the designated subject and object out of all objects within that region. 

However, the use of raw subgraph features fails to capture the interactions between the subgraph and objects within it. To resolve this problem, \emph{Context} module from \cite{li2018factorizable} is modified and adopted. It is designed based on the attention mechanism \cite{vaswani2017attention}. It first guides the model to ``look at" the potential locations where subject and object are located, and then aggregate subgraph features using the object features from the corresponding spatial location. Meanwhile, since different objects exist at different spatial locations, weights should be spatial-dependent when incorporating object features into subgraph features.

The detailed description is as follows. Let there be $K$ objects within a subgraph. The weighted object features at location $(x, y)$ are expressed as below:
\begin{equation}
    \hat{\textbf{O}}(x, y) = \sum_{i=1}^{K}W(\textbf{o}_i, S, x ,y) \cdot V_{proj}(\textbf{o}_i)
\end{equation}
where $S$ denotes the subgraph features. $W(\textbf{o}_i, S, x, y)$ is the attention weight of object feature $\textbf{o}_i$ at $(x, y)$, and $\hat{\textbf{O}}(x, y)$ represents the weighted object features at the same location. $V_{proj}$ projects $\textbf{o}_i$ to the target subgraph domain. The attention weight is computed as: 

\begin{small}
\begin{equation}
    W(\textbf{o}_i, S, x, y) = \frac{exp(K_{proj}(\textbf{o}_i) \cdot Q_{proj}(S(x, y)))}{\sum_{i=1}^{N}exp(K_{proj}(\textbf{o}_i) \cdot Q_{proj}(S(x, y)))}
\end{equation}
\end{small}\\
where $S(x, y)$ denotes the subgraph feature at $(x, y)$.  $K_{proj}$ and $Q_{proj}$ are projection functions. Then we represent the aggregated subgraph features as:

\begin{equation}
    \hat{S} = \sum_{x, y}S(x, y) + \alpha \cdot \hat{\textbf{O}}(x, y)
\end{equation}
where $\alpha$ is a learnable scale factor. 

\subsection{Manipulation Relationship Graph Construction}
After obtaining the refined object and subgraph features, a manipulation relationship is predicted using \emph{\textbf{AR-Net}} by taking the addition of the corresponding subject, object and subgraph features as follows:
\begin{equation}
    p_{ij} = FC(ReLu(\textbf{o}_{i} + \hat{S} +\textbf{o}_{j}))
\end{equation}
where $p_{ij}$ denotes the relationship (from $i$ to $j$) distribution over $R$ classes of manipulation relationships plus background (no relationship). $FC$ is the fully-connected layer and $ReLu$ is the activation function.

Top-1 prediction will be used to form the manipulation relationship triplet as ($subject$, $relationship$, $object$), where subject and object are represented by their indexes. Triplet score is computed by taking the product of subject, object and relationship confidence probabilities. Redundant triplets and those with low scores will be eliminated with triplet NMS \cite{li2017vip}. After that, an MRG is constructed by treating objects as nodes and relationships between pairs of objects as edges with triplet scores being edge weights. 

Overall, the proposed \emph{\textbf{AR-Net}} uses VGG16 \cite{simonyan2014very} pretrained on ImageNet \cite{krizhevsky2012imagenet} as backbone and Adam \cite{kingma2014adam} as optimizer with 0.9 momentum and 0.00001 weight decay. Initial learning rate is set as 0.001 and get multiplied by 0.1 every 5 epochs. Subgraph clustering threshold is set to 0.5. RPN is trained first and then the whole model is trained jointly for 15 epochs. The model is designed based on \cite{ren2015faster} \cite{li2018factorizable}.

\section{Experimental Setup} \label{exp_setup}
In this section, we will first discuss the details of a novel manipulation relationship dataset that is used to evaluate the proposed \emph{\textbf{AR-Net}} and other baselines. The setup for both visual perception and physical manipulation experiments will then be presented. Finally, the evaluation metrics used for experiments will be given. 

\subsection{SMRD Dataset}

\begin{table}[h]
\centering
\caption{Summary of SMRD dataset}
\renewcommand{\arraystretch}{1.3}
\begin{tabular}{|c|c|}
\hline
Relationship class & scoop, pour, cut, contain, wipe, dump                                                                                        \\ \hline
Tool class         & \begin{tabular}[c]{@{}c@{}}pan, spatula, plate, knife,  bowl, cloth, \\ fork, mug,  spoon, brush, cup, pot, can\end{tabular} \\ \hline

Image split        & \begin{tabular}[c]{@{}c@{}}988 for training, 183 for testing, \\ 1171 images in total\end{tabular}                                  \\ \hline
Relationship split & \begin{tabular}[c]{@{}c@{}}3563 for training, 865 for testing, \\ 4428 relationships in total\end{tabular}                    \\ \hline
\end{tabular}
\label{tab:dataset_summary}
\end{table}

\begin{figure}[t]
  \centering
  \begin{tikzpicture}[inner sep = 0pt, outer sep = 0pt]
    \node[anchor=south west] (fnC) at (0in,0in)
      {\includegraphics[height=2in,clip=true,trim=0in 0in 0in 0in]{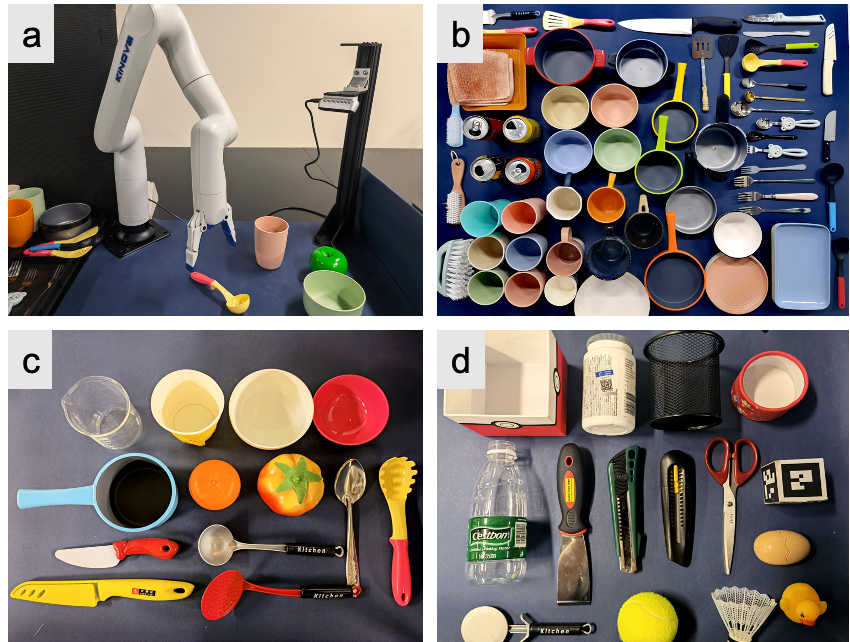}};
  \end{tikzpicture}
  \vspace*{-0.1in}
  \caption{(a) Experiment platform; (b) Kitchen tools used in SMRD dataset; (c) Intra-class objects; (d) Inter-class objects.} 
  \label{fig:exp_setup}
  \vspace*{-0.25in}
\end{figure}

In order for a robot to learn diverse manipulation relationships, a dataset annotated with ground truth manipulation relationships is needed. Not aware of such a dataset, Semantic Manipulation Relationship Detection (SMRD) dataset is created. A summary of SMRD is presented in Table \ref{tab:dataset_summary}. It includes 13 kitchen tool classes as shown in Figure \ref{fig:exp_setup} (b) and six manipulation relationship classes. Non-affordance objects (e.g. vegetables, fruits) are also collected since manipulation relationships commonly exist between tools with specific affordances and non-affordance objects (e.g. ($knife$, $cut$, $apple$)). Overall, SMRD consists of 1171 images and 4428 relationships, averaging 3.78 relationships per image. For each image, the GT relationships are organized in the form of an MRG. The dataset does not consider the category label since the label is not used during relationship inference.

\subsection{Visual Perception Experiment}
To gain insights into our proposed model and show the effectiveness of each component, visual perception experiment is conducted. We compare \emph{\textbf{AR-Net}} to three baselines:\\
\begin{itemize}
    \item[$\bullet$] \emph{\textbf{Base}}
    uses raw object and subgraph features. It does not consider the attribute of each object and isolate the contextual connection between relationship region and objects within it. 
    
    \item[$\bullet$] \emph{\textbf{Base+ATT}}
    refines object features with \emph{Attribute} module. It guides the relationship learning by explicitly predicting the attributes of objects. 
    
    \item[$\bullet$] \emph{\textbf{Base+CT}}
    is equipped with \emph{Context} module. It aggregates subgraph features with object features from the corresponding spatial location and therefore guides the model to ``look at" the positions where subject and object are located.
    
    \item[$\bullet$] \emph{\textbf{Base+ATT+CT}} (\emph{\textbf{AR-Net}}) takes advantage of both \emph{Attribute} module and \emph{Context} module. It incorporates attribute learning and context information to obtain better transferability and prediction accuracy.
\end{itemize}

\subsection{Physical Manipulation Experiment}
We then evaluate the feasibility of the proposed method in robotic manipulation experiment. Figure \ref{fig:exp_setup} (a) shows the experiment platform. It consists of a 6-DOF Kinova Gen3 Lite robotic arm and an Intel Realsense D435i RGBD camera with eye-to-hand calibration. In addition to  \emph{\textbf{AR-Net}}, two types of affordance learning approaches are presented for comparison:
\begin{itemize}
    \item[$\bullet$] \emph{\textbf{AS}}
    represents segmentation-based affordance learning methods \cite{myers2015affordance, do2018affordancenet, xu2021affordance, chu2019recognizing, deng20213d}. In these methods, object parts sharing the same functionality are segmented and grouped at the pixel level. In the following comparison, we select state-of-the-art affordance segmentation architecture \cite{do2018affordancenet} for comparison.
    
    \item[$\bullet$] \emph{\textbf{DKG}} first detects the object categories in the given scene and then query the trained knowledge base to infer the manipulation relationships.  \cite{saxena2014robobrain} is reproduced as a baseline approach. It consists of an object detector pretrained on SMRD in conjunction with a knowledge graph constructed from GT manipulation relationships for fair comparison.
\end{itemize}

\begin{figure}[t]
  \centering
  \begin{tikzpicture}[inner sep = 0pt, outer sep = 0pt]
    \node[anchor=south west] (fnC) at (0in,0in)
      {\includegraphics[height=1in,clip=true,trim=0in 0in 0in 0in]{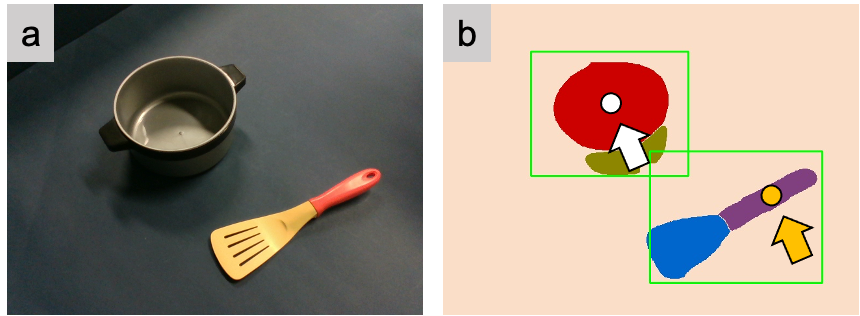}};
  \end{tikzpicture}
  \vspace*{-0.1in}
  \caption{Grasp point (yellow) and effect point (white) computed using AffordanceNet \cite{do2018affordancenet}} 
  \label{fig:grasp_point}
    \vspace*{-0.25in}
\end{figure}

Since each method above deals with a unique set of affordance categories, four shared affordances  including ``scoop", ``pour", ``cut", ``contain" are extracted for testing. Besides, for the purpose of robotic manipulation, grasp point on subject and effect point on object are computed using AffordanceNet \cite{do2018affordancenet}. It is done by first performing affordance segmentation on subject and object. Grasp and effect points are then computed by taking the midpoint of selected affordance regions. As is shown in Figure \ref{fig:grasp_point},  AffordanceNet segments the pot to a ``contain" region (red) and a ``wrap-grasp" region (brown). Spatula is segmented into a ``scoop" region (blue) and a ``grasp" region (purple). Grasp point (yellow) is the midpoint of ``grasp" region on spatula (subject) while effect point (white) is the midpoint of ``contain" region on pot (object). Finally, manipulation action is executed using motion primitives.

\subsection{Evaluation Metric}
Two sets of evaluation metrics are provided for visual perception and physical manipulation experiments respectively. For the former, Recall@1 (denoted as R@1) and Recall@5 (denoted as R@5) are used. 
We evaluate the performance under the tasks of phrase detection (denoted as P) and relationship detection (denoted as R). The former task coarsely localizes the entire relationship as one union bounding box having at least 0.5 overlap with the ground truth box and the latter requires the localization of both the subject and the object with respect to their ground truth bounding boxes and predicts the correct relationship simultaneously. More details can be found in \cite{lu2016visual}. In all, four metrics including PR@1, PR@5, RR@1, RR@5 are used in visual perception experiment. For physical manipulation experiment, the robot needs to localize each object precisely for the purpose of manipulation. Therefore RR@1 and RR@3 are used. To test the task execution capability, task relationship recognition rate (denoted as TRR) and task completion rate (denoted as TC) are recorded. TRR examines robot's recognition ability towards a given task relationship and TC evaluates the task completion capability.

\section{Experiments} \label{exp}
This  section  details  the  experiments  performed  and  their outcomes. We first show the results from visual perception experiment by comparing the proposed method to three baselines. Three sets of physical manipulation experiments are then conducted on the seen, intra-class and inter-class objects respectively.

\subsection{Visual perception experiment results}

\begin{table}[h]
\centering
\caption{performance on visual experiments (\%)}
\renewcommand{\arraystretch}{1.3}
\label{tab:visual_exp}
\begin{tabular}{|c|c|c|c|c|}
\hline
Method    & PR@1 & PR@5 & RR@1 & RR@5 \\ \hline
Base& 28.704     & 74.884     & 17.130     & 59.375     \\ \hline
Base+ATT  & 35.764     & 82.060     & 19.792     & 67.593     \\ \hline
Base+CT   & 35.995     & 81.597     & 19.907     & 67.246     \\ \hline
\emph{\textbf{Base+ATT+CT (proposed)}} & \textbf{37.384}     & \textbf{84.259 }    & \textbf{20.370 }    & \textbf{68.059}     \\ \hline
\end{tabular}
\end{table}

Visual perception experiment is performed on SMRD. Four models follow the same training and inference strategies. As shown in Table \ref{tab:visual_exp}, the proposed \emph{\textbf{Base+ATT+CT}} outperforms all three baselines across all four metrics. This highlights the model's ability to accurately recognize manipulation relationships using attribute and context information. \emph{\textbf{Base}} shows the lowest recall since it does not utilize object attribute information and does isolate the connection between relationship region and objects within it. \emph{\textbf{Base+ATT}} gains a performance boost compared to \emph{\textbf{Base}}. It shows the effectiveness of proposed \emph{Attribute} module. Instead of predicting the category label, it guides the feature learning by inferring what attributes an object possesses. \emph{\textbf{Base+CT}} also shows competitive performance as it bridges the gap between relationship region and objects within it. With \emph{Context} module, the model learns both ``where to look at" (spatial info) and ``what to look at" (appearance info).

\begin{figure}[t]
  \centering
  \begin{tikzpicture}[inner sep = 0pt, outer sep = 0pt]
    \node[anchor=south west] (fnC) at (0in,0in)
      {\includegraphics[height=2in,clip=true,trim=0in 0.2in 0in 0in]{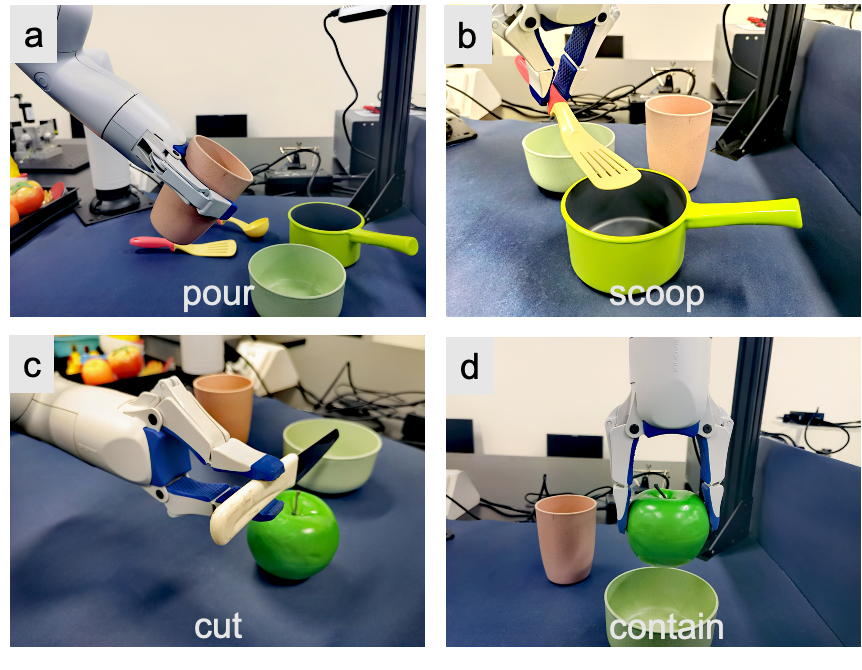}};
  \end{tikzpicture}
  \vspace*{-0.1in}
  \caption{Examples of manipulation relationships executed by the robot} 
  \label{fig:4task}
\end{figure}

\subsection{MRG guided task execution}

\begin{table}[h]
\centering
\caption{In-class task execution (\%)}
\renewcommand{\arraystretch}{1.3}
\label{tab:MRG_task}
\begin{tabular}{|c|c|c|c|c|}
\hline
Method      & RR@1  & RR@3  & TRR   & TC    \\ \hline
AS       & 23.73 & 45.41 & 33.33 & 33.33 \\ \hline
DKG        & \textbf{29.29} & 78.41 & 86.67 & 66.67 \\ \hline
\emph{\textbf{AR-Net (Ours) }}& 29.26 & \textbf{80.64} & \textbf{93.33} & \textbf{80.00} \\ \hline
\end{tabular}
\end{table}

For the following three physical tasks, we examine each method's ability to recall the manipulation relationships and physically manipulate objects with motion primitives. Shown in Figure \ref{fig:4task} are examples of manipulation relationships executed by the robot. Seen objects from SMRD are first tested.

In this task, a set of objects covering all four manipulation relationships (i.e. scoop, pour, cut, contain) are selected from the SMRD and 15 scenes are constructed. Each scene consists of 3-7 relationships, and one task relationship is randomly selected from GT relationships for physical execution. The robot is required to (1) recall all the manipulation relationships within the given scene (RR@k), (2) recognize the selected task relationship (TRR), (3) physically execute the task with motion primitives (TC). Results are presented in Table \ref{tab:MRG_task}. \emph{\textbf{AS}} performs the worst. It is because segmentation-based approach only deals with objects with specific affordances such as tools whereas most manipulation actions happen between tools and non-affordance objects. For example, \emph{\textbf{AS}} fails to recognize and execute the task ($knife$, $cut$, $apple$) since apple has no affordance. Therefore, it fails in all ``cut" and ``contain" tasks. \emph{\textbf{DKG}} gives competitive result, leading in RR@1. The model is able to output robust detection results and query the knowledge base accordingly. However, the misclassifications lead to inferior performance to the proposed method across three metrics as \emph{\textbf{AR-Net}} does not rely on object categories to make predictions.

\subsection{Intra-class generalization}

\begin{table}[h]
\centering
\caption{Intra-class generalization (\%)}
\renewcommand{\arraystretch}{1.3}
\label{tab:intra_class}
\begin{tabular}{|c|c|c|c|c|}
\hline
Method      & RR@1  & RR@3  & TRR   & TC    \\ \hline
AS        & 18.50 & 44.17 & 26.67 & 26.67 \\ \hline
DKG        & \textbf{26.83} & 68.33 & 86.67 & 73.33 \\ \hline
\emph{\textbf{AR-Net (Ours) }}& 26.28 & \textbf{72.72} & \textbf{93.33} & \textbf{80.00} \\ \hline
\end{tabular}
\end{table}

In this task, we investigate whether each model can generalize the learned relationships to unseen objects from the same class. 

As shown in Figure \ref{fig:exp_setup} (c), a set of objects sharing the same category with SMRD is collected but they vary in color, material and size. Table \ref{tab:intra_class} details the results. Similarly, \emph{\textbf{AS}} gives the lowest recall and task completion rate. \emph{\textbf{DKG}} serves as a strong baseline and performs slightly better than \emph{\textbf{AR-Net}} in terms of RR@1. A drop occurs on RR@3 for both \emph{\textbf{DKG}} and \emph{\textbf{AR-Net}}, showing the large intra-class variation. The proposed method achieves superior RR@3, TRR and TC due to the category-agnostic design, leading to better task execution ability.

\subsection{Inter-class generalization}

\begin{table}[h]
\centering
\vspace{-0.1in}
\caption{Inter-class generalization (\%)}
\renewcommand{\arraystretch}{1.3}
\label{tab:inter_class}
\begin{tabular}{|c|c|c|c|c|}
\hline
Method& RR@1  & RR@3  & TRR   & TC    \\ \hline
AS       & 8.73  & 13.02 & 20.00 & 20.00 \\ \hline
DKG        & 27.06 & 51.35 & 66.67 & 46.67 \\ \hline
\emph{\textbf{AR-Net (Ours)}} & \textbf{28.17} & \textbf{62.30} & \textbf{80.00} & \textbf{60.00} \\ \hline
\end{tabular}
\vspace{-0.1in}
\end{table}

Figure \ref{fig:exp_setup} (d) shows the objects used for inter-class generalization experiment, which is the most challenging task out of three. These objects differs largely from SMRD in usage, appearance and material. Results are shown in Table \ref{tab:inter_class}. The proposed method outperforms all other baselines across all four metrics by significant margins. Large inter-class variation causes considerable misclassifications to \emph{\textbf{DKG}} and therefore leads to inferior performance. \emph{\textbf{As}} performs the worst as the model does not generalize well to inter-class objects and meanwhile is not able to recognize non-affordance objects.

\begin{figure}[t]
  \centering
  \begin{tikzpicture}[inner sep = 0pt, outer sep = 0pt]
    \node[anchor=south west] (fnC) at (0in,0in)
      {\includegraphics[height=2in,clip=true,trim=0in 0in 0in 0in]{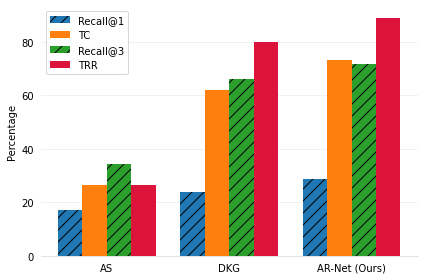}};
  \end{tikzpicture}
    \vspace*{-0.1in}
  \caption{Overall performance (\%) of three methods in physical experiments} 
  \label{fig:overall_exp}
  \vspace*{-0.25in}
\end{figure}

An overall evaluation of three methods is presented in Figure \ref{fig:overall_exp} by combining the statistics from three sets of physical experiments. The proposed method achieves the success rate of 88.89\% on TRR and 73.33\% on TC, showing clear superiority under each evaluation metric.

\section{Discussion and Conclusion} \label{discussion}

\begin{figure}[h]
  \centering
  \vspace*{-0.1in}
  \begin{tikzpicture}[inner sep = 0pt, outer sep = 0pt]
    \node[anchor=south west] (fnC) at (0in,0in)
      {\includegraphics[height=2in,clip=true,trim=0in 0in 0in 0in]{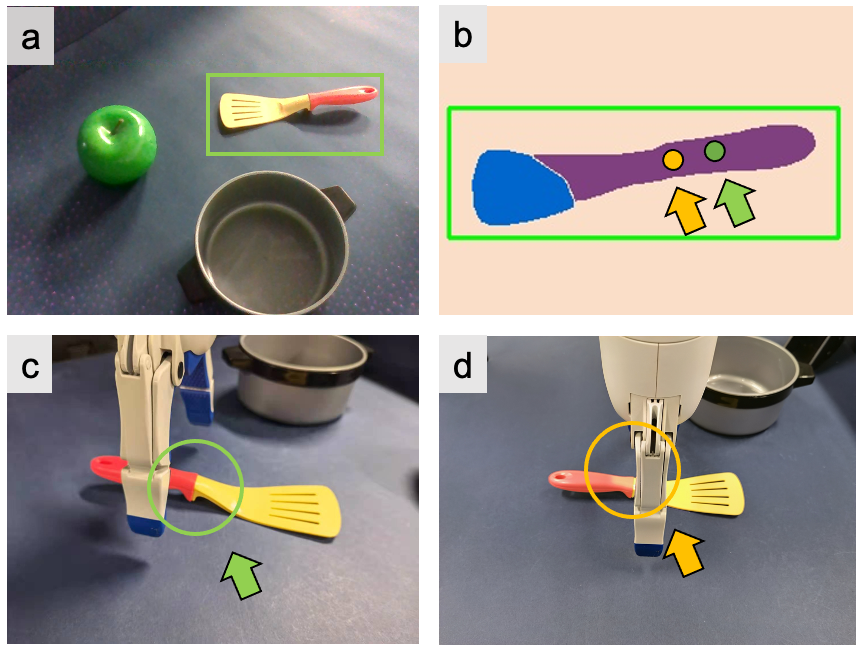}};
  \end{tikzpicture}
      \vspace*{-0.1in}
  \caption{Human annotated (GT) grasp point (green) and computed grasp point (yellow)} 
  \label{fig:error_analysis}
  \vspace*{-0.12in}
\end{figure}

A significant drop from TRR to TC is observed from Figure \ref{fig:overall_exp}, showing the gap between relationship recognition and successful manipulation. The drop is mainly caused by two reasons. First, a manipulation action is implemented by way of pre-designed motion primitives, which do not consider the task and environment constraints. Low-level task and motion planning (TAMP) algorithm can be combined to generate flexible manipulation actions. Another typical failure case is due to sub-optimal grasp points. As is shown in Figure \ref{fig:error_analysis}, grasp point computed using \cite{do2018affordancenet} drifts away from the GT grasp point annotated by human, leading to sub-optimal grasp poses and potential manipulation failure. This indicates that task-oriented grasping (TOG) module \cite{qin2020keto} \cite{liu2020cage} is needed.

To conclude, this work addresses the problem of affordance learning in the task of robotic manipulation. We introduce a novel affordance representation which captures the potential manipulation relationships of an arbitrary scene. A deep neural network is designed to construct the proposed MRG from raw visual observations. To train the model, a novel dataset is collected and annotated, consisting of six manipulation relationship classes and 13 tool categories. Visual perception and physical manipulation experiments demonstrate the superiority of proposed method and representation against all baselines. For the next step, TAMP and TOG modules will be incorporated to generate task-oriented manipulation trajectories and grasp poses in order to enhance the physical manipulation capability.




\section*{ACKNOWLEDGMENT}

We thank members of Robotics Perception and Intelligence Lab at SUSTech for their helpful feedback and discussion.


\bibliographystyle{IEEEtran}
\balance
\bibliography{main}

\end{document}